\documentclass[a4paper,conference]{IEEEtran}

\usepackage{graphicx} % Enhanced LaTeX Graphics
\usepackage{svg}
\usepackage{subfigure} % subcaptions for subfigures
\usepackage{subfigmat} % matrices of similar subfigures
\usepackage[left=1.57cm,right=1.57cm,top=0.95cm,bottom=2.54cm]{geometry}

\usepackage{dcolumn}           % decimal-aligned tabular math columns
\newcolumntype{.}   {D{.}{.}{-1}} % column alignedd on the point separator '.'
\newcolumntype{d}[1]{D{.}{.}{#1}} % column centered on the point separator '.'
\newcolumntype{e}   {D{E}{E}{-1}} % column centered on the exponent 'E'
\newcolumntype{E}[1]{D{E}{E}{#1}} % column centered on the exponent 'E'

\usepackage{amsmath}
\usepackage{amsfonts}
\usepackage{amssymb}
\usepackage{amsthm}

\usepackage{float}
\usepackage[space]{cite}

\usepackage{url}
\usepackage{verbatim}

\usepackage{titlesec}

\titlespacing*{\section}{0pt}{10pt}{0pt}
\titlespacing*{\subsection}{0pt}{10pt}{0pt}

\usepackage{secdot}
\sectiondot{subsection}
\sectionpunct{section}{. }    % By default, \sectiondot places a \quad
\sectionpunct{subsection}{. } % after the number

\DeclareMathOperator*{\argmax}{arg\,max}

\title{Assessing Policy, Loss and Planning Combinations in Reinforcement Learning using a New Modular Architecture}
\date{October 2021}
\author{\IEEEauthorblockN{Tiago Gaspar Oliveira}
\IEEEauthorblockA{\textit{INESC-ID / Instituto Superior Técnico} \\
Lisbon, Portugal \\
tiagojgroliveira@tecnico.ulisboa.pt}
\and
\IEEEauthorblockN{Arlindo L. Oliveira}
\IEEEauthorblockA{\textit{INESC-ID / Instituto Superior Técnico} \\
Lisbon, Portugal \\
arlindo.oliveira@tecnico.ulisboa.pt}
}

\begin{document}

% Begin one column section for title and abstract
%
% http://www.faqs.org/faqs/de-tex-faq/part5/
%\twocolumn[
%\begin{@twocolumnfalse}
\maketitle

%%%%%%%%%%%%%%%%%%%%%%%%%%%%%%%%%%%%%%%%%%%%%%%%%%%%%%%%%%%%%%%%%%%%%%
% ABSTRACT & KEYWORDS
%%%%%%%%%%%%%%%%%%%%%%%%%%%%%%%%%%%%%%%%%%%%%%%%%%%%%%%%%%%%%%%%%%%%%%

\begin{abstract}
The model-based reinforcement learning paradigm, which uses planning algorithms and neural network models, has recently achieved unprecedented results in diverse applications, leading to what is now known as deep reinforcement learning. These agents are quite complex and involve multiple components, 
factors that can create challenges for research. In this work, we propose a new modular software architecture
suited for these types of agents, and a set of building blocks that can be easily reused and assembled to construct new model-based reinforcement learning agents. These building blocks include planning algorithms, policies, and loss functions
%\footnote{Reusable source code will be made available, once anonymity requirements are dropped upon acceptance of this paper.} 
\footnote{A modeular Python3 implementation is available at the location \url{https://github.com/GaspTO/Modular\_MBRL}}.

We illustrate the use of this architecture by combining several of these building blocks to implement and test agents that are optimized to three different test environments: Cartpole, Minigrid, and Tictactoe. One particular planning algorithm, made available in our implementation and not previously used in reinforcement learning, which we called \textit{averaged minimax}, achieved good results in the three tested environments. 

Experiments performed with this architecture have shown that the best combination of planning algorithm, policy, and loss function is heavily problem dependent. This result provides evidence that the proposed architecture, which is modular and reusable, is useful for reinforcement learning researchers who want to study new environments and techniques.
%%
%% Keywords (max 5)
%%

\vspace{0.2cm}
\end{abstract}

% End one column section (begin default two columns)
%%\end{@twocolumnfalse}
%%]
%%%%%%%%%%%%%%%%%%%%%%%%%%%%%%%%%%%%%%%%%%%%%%%%%%%%%%%%%%%%%%%%%%%%%%
% INTRODUCTION
%%%%%%%%%%%%%%%%%%%%%%%%%%%%%%%%%%%%%%%%%%%%%%%%%%%%%%%%%%%%%%%%%%%%%%

\section{Introduction}
\label{sec:intro}
In 2016, a program called AlphaGo \cite{AlphaGo} beat a Go world champion for the first time. Go was seen as the new milestone of artificial intelligence since the former world chess champion, Gary Kasparov, lost to DeepBlue in 1997 \cite{DeepBlue}. Go was finally mastered using reinforcement learning (RL), neural networks and a search algorithm. Following this achievement, an even more powerful program called AlphaGo Zero \cite{AlphaGoZero} was created, learning exclusively from playing against itself, indicating the strong potential of deep reinforcement learning (DRL).\par
It is not that any of those elements brought anything completely new to the field, but it was the first time they were assembled together and achieved outstanding results. The search component allowed deep reinforcement learning to reach a completely new level. Following this line of algorithms, two other very successful agents were created: AlphaZero, a generalized version of AlphaGo Zero, that became the best player in chess, shogi and Go; and MuZero, a program that achieved the same outstanding results, but was not given even the rules of the game, which were learned as it played. Learning without being given the structure of the state space is, of course, very important  since most of real-world problem dynamics do not have a clear set of rules that can be explained to a computer beforehand. \par
Despite all of this, there remains an overwhelming barrier concerning the more extensive use of these algorithms in a wider range of problems. The computational resources used to train these models were much higher than what is usually available to researchers (AlphaZero used more than 5000 first-generation Tensor Processing Units \cite{AlphaZero}). It is only normal that, after their success, the research on these algorithms has been increasing, trying to make them more practical to use \cite{COnAndOffPolicyMuzero,TwoPlayerInfo,ConstrainHiddenSpace,SAVE,AlternativeLossFunctions}. \par
State of the art reinforcement learning algorithms, especially these new model-based algorithms that use search, are complex and involve multiple interacting parts. Implementing these algorithms can be very time consuming, one of the reasons being deep learning's proclivity to \textit{fail silently} due to its adaptable nature, which often leads to a large amount of time spent \textit{debugging} and verifying code. It is not simply practical to refactor and heavily recode the implementation every time one wants to try and study different ideas. And, even though there has been a recent effort to publish open source implementations, these do not usually take their possible extensions into account and are hard to addapt. Our work attempts to help with these problems by introducing a modular software architecture for model-based reinforcement learning (MBRL). Its objective is to separate the agent into distinctive components and facilitate the implementation of new strategies (the algorithms in each one of them), that can be added without having to modify the other components. The architecture should enable us to construct multiple agents easily by choosing different strategy combinations. \par 
We provide several different instantiations of the modules of this architecture, using some common solutions for planning, policies and loss functions. One of the instances made available, namely a new simple planning algorithm called averaged minimax, performed well in the three environments tested and may be competitive in other environments. We also present a comparative study of the agents created by combining different modules and show that the best agent depends heavily on the nature of the specific problem being addressed.

%%%%%%%%%%%%%%%%%%%%%%%%%%%%%%%%%%%%%%%%%%%%%%%%%%%%%%%%%%%%%%%%%%%%%%
% BACKGROUND
%%%%%%%%%%%%%%%%%%%%%%%%%%%%%%%%%%%%%%%%%%%%%%%%%%%%%%%%%%%%%%%%%%%%%%

\section{Background}
\label{sec:backg}

\subsection{Reinforcement Learning}
In reinforcement learning, an agent interacts with an environment, that is usually described as a Markov decision process (MDP). At each time step $t$, the agent will decide the next action, $a_t$, to take. The environment will receive this action and transition its state, $s_t$, to the next one, $s_{t+1}$, returning an associated reward $r_t$. The agent continues to interact with the environment until the episode reaches a terminal state, originating a sequence of transitions, called \textit{trajectory}, $s_0, a_0, r_0, s_1, a_1, r_1, s_2 ...$ \par 
These environments have the Markov property, meaning that the result (next state and reward) after an action has been taken depends solely on the current state and not on the prior history of the current episode. In an MDP, there is a state space $\mathcal{S}$, an action space $\mathcal{A}$, a probability function  $p(s'|s,a)$ that calculates the likelihood of the agent ending in state $s'$ if action $a$ is chosen in state $s$,  and a reward function  $r(s,a,s')$  that gives the reward associated with this transitions. These two functions describe~ the dynamics of the environment, for all $s', s \in \mathcal{S}$ and $a \in \mathcal{A}$. \par
There are also certain environments that have the Markov property but where states are not completely visible to the agent. These environments are formulated as a partially observable Markov decision processes (POMDPs). In these, an agent has to act under the uncertainty of its current state, only having access to an observation. MDPs are just special cases of POMDPs.\par

\subsection{Muzero}
Muzero is a model based reinforcement learning (MBRL) agent. It uses Monte Carlo tree search (MCTS) \cite{MCTS} to augment the capacity of its state evaluation by looking ahead in the state-space. In each iteration, the search will descend the tree looking for a leaf node to expand, using a best-first formula that mediates between exploitation and exploration. When expanding, it estimates a value for the leaf node's state, which influences and estimates a policy distribution for the successors. The value estimated is added to the total value of each node in the path descended and the average of these values makes up the improved state value of that node, $V(s)$. The search repeats for a certain number of iterations and, in the end, when it is time to choose an action to use in the \textit{real} environment, the successors of the root node define a policy function, $\pi$, using their visit distribution:
\begin{equation}
	\label{AlphaZeroP}
	\pi(a|s) = \frac{N(s,a)^{1/\tau}}{\sum_{b}N(s,b)^{1/\tau}}  ,
\end{equation}
where $N(s,a)$ is the number of visits associated with the transition $(s,a)$ and $\tau$ is a \textit{temperature parameter}, influencing the exploration/exploitation trade-off. The lower it is, the greedier the policy becomes. \par 
Muzero learns the environment by interacting with it, and the states learned do not have specific semantics. The environment is learned through a representation function, $h_\theta$, that converts a real observation to a hidden state and a dynamics function, $g_\theta$, that receives a hidden state and an action and returns the next state and reward associated. The policy and value are estimated by a prediction function, $f_\theta$, which receives a hidden state as argument. An observation from a played episode is sampled and is unrolled by applying the representation function, converting it to an hidden state, followed by applying the next $k$ actions used in the episode to the successive hidden states, using the dynamics function. Then, the prediction function is used in each of these hidden states, returning the policy and value estimations. To learn these functions, the rewards obtained are updated to match the observed ones. The policy is updated based on the visit distribution given by the root node of the respective observation and the value function is improved based on a Monte Carlo value or by bootstrapping from an improved state value.

%%%%%%%%%%%%%%%%%%%%%%%%%%%%%%%%%%%%%%%%%%%%%%%%%%%%%%%%%%%%%%%%%%%%%%
% ARCHITECTURE
%%%%%%%%%%%%%%%%%%%%%%%%%%%%%%%%%%%%%%%%%%%%%%%%%%%%%%%%%%%%%%%%%%%%%%

\section{Architecture}
\label{sec:architecture}
\begin{figure}[h]
	\centering
	%\includesvg[width=0.85\linewidth]{./images/architecture_svg.svg}
	\includegraphics[width=0.85\linewidth]{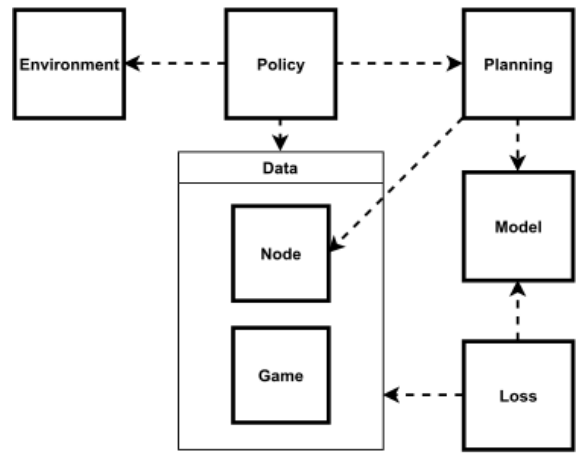}
	\caption{Basic architecture overview.}
	\label{fig:architecture}
\end{figure}

AlphaGo Zero, AlphaZero, Muzero and other variations of these algorithms \cite{TwoPlayerInfo,SAVE,ThinkingFastAndSlow} share a common structure and behavior that we generalize into an architecture. This architecture has the objective of facilitating the process of making changes and extensions in each key component, limiting the complexity by using modularity. The reasoning behind it is to identify key parts in these agents and implement each of these parts as independently as possible, delegating the responsibility to the user to assemble each block consistently, according to the intended agent. We propose an architecture with six different parts, which are called modules: Environment, Loss, Model, Planning, Policy and Data. We consider every one of these components, except the Environment itself, to correspond to the concept of an agent. The Loss, Model, Planning and Policy modules determine how the agent acts and learns, while the Data module concerns itself with the storage of data to be used by the other four modules. The purpose of each is to have alternative algorithms that can be chosen to accomplish the module’s objective. The architecture was conceived to make it easy to add a new algorithm to each module, without having to refactor or recode anything else. A simple scheme summarizing the structure of the architecture is given in figure \ref{fig:architecture}, where the arrows show dependencies between modules. \par

The agent interacts with an \textbf{environment} through a \textbf{policy}. In each step of the episode, the policy will call a \textbf{planning} algorithm to \textit{look ahead} into the simulated state-space. After the planning algorithm is done, it returns the root node (that coincides with the current state) back to the policy. Based on the attributes of this root node, the policy decides the next action to take, advancing to the next state. The policy saves the nodes and the data returned by its interactions with the environment using the \textbf{Data} module. Then, this stored data can be passed to a \textbf{loss} algorithm that calculates a loss value in order to update the model. Lastly, the architecture establishes the relationship between these blocks, but the user is the one who has to assemble them. It has to instantiate all the specific algorithms and then create the agent’s logic, which defines when to call the policy is to receive data and when to call the loss to update the model.

%%%%%%%%%%%%%%%%%%%%%%%%%%%%%%%%%%%%%%%%%%%%%%%%%%%%%%%%%%%%%%%%%%%%%%
\subsection{Environment}
A RL agent interacts with a problem, which we call environment. The agent needs to know how to interact with the environment’s interface, but not vice-versa. We assume they are implemented as POMDPs.

%%%%%%%%%%%%%%%%%%%%%%%%%%%%%%%%%%%%%%%%%%%%%%%%%%%%%%%%%%%%%%%%%%%%%%
\subsection{Model}
This module’s function is to model not only the environment, but also all the other functions required by the Planning module. Different search algorithms might need new functions, so it is important for this component’s implementation to be efficient and easily extendable. In this module, for the functions we want to learn, we assume them to be computed using neural networks, but do not make assumptions about their internal structure. The architecture should support both MBRL agents that learn the environment’s model and those that receive a perfect one. In case this module is queried for the next state after an action, if it knows the environment’s dynamics, it can return the true next state. If it does not, it returns an estimation.

%%%%%%%%%%%%%%%%%%%%%%%%%%%%%%%%%%%%%%%%%%%%%%%%%%%%%%%%%%%%%%%%%%%%%%
\subsection{Data}
The function of the Data module is to store and transport information between modules. We divide it into two submodules. The first is the Node submodule. A node coincides conceptually with an environment state and it is created by the planning algorithm to form the search graph/tree. The particular node class used is dependent on each search algorithm. \par
The second is the Game submodule and it stores all the data seen during the episode, both by the environment (e.g. observations) and the search trees returned by the planning.

%%%%%%%%%%%%%%%%%%%%%%%%%%%%%%%%%%%%%%%%%%%%%%%%%%%%%%%%%%%%%%%%%%%%%%
\subsection{Planning}
We consider state-space planning algorithms. In these, actions mediate transitions between two states. In traditional DRL algorithms the agent estimates, at each step, the quality of the possible actions available using a neural network. The objective of the Planning module is to provide enough information about each action, in a given step of the episode, so the policy can choose the best possible action. However, it does this differently from traditional DRL algorithms. Instead of using a neural network alone to estimate the quality of each action, it uses a search algorithm in conjunction with the functions available in the model passed. The reasoning behind using a search algorithm is to look ahead into the future of the space of (hypothetical) states in order to provide more accurate quality estimations about each action. This module does not come up with a plan to use during the actual interaction with the environment. In fact, it should not even be aware of the existence of an episode. It simply makes estimations just like a normal neural model would in a typical DRL agent. The root node, which represents the current state of the environment, and has attached to it the rest of the tree,  will be returned to the policy. Lastly, as mentioned, the planning class should receive a model. How the model works internally is irrelevant, as long as it can calculate all the required functions.

%%%%%%%%%%%%%%%%%%%%%%%%%%%%%%%%%%%%%%%%%%%%%%%%%%%%%%%%%%%%%%%%%%%%%%
\subsection{Policy}
The Policy module has two responsibilities. It decides the action to take in every step of the episode, and it stores the relevant data coming from the environment and planning in the Data module. The policy iterates through the whole environment episode, balancing between exploration and exploitation. In every step, it calls a search algorithm receiving the root node corresponding to the current state. Then chooses which action to take. It is also responsible for storing the whole episode in an object of the Game submodule. In the end, after completing an episode, it will save in an object of the Game Module all the information returned by the environment (observations, rewards…) and the nodes returned by the planning.  \par
Each policy should receive the instance of the planning class to use. As long as the node returned by it has the interface needed by the policy, every planning algorithm should work. For instance, if the policy wants to use the number of times an action has been chosen during search (as in AlphaZero), then any search algorithm that returns a node with this information (which is usually associated with best-first algorithms) should work. The user should call the policy when it wants to acquire more data (in the form of Game instances).

%%%%%%%%%%%%%%%%%%%%%%%%%%%%%%%%%%%%%%%%%%%%%%%%%%%%%%%%%%%%%%%%%%%%%%
\subsection{Loss}
After interacting with the environment, the model should be updated to be more accurate and insightful. This is done by using the experience previously returned by the policy, which is passed to a loss algorithm, returning a loss value, a computational graph, that can then be used to update the neural functions.  Each specific loss algorithm is dependent on the particular characteristics of the problem, so it does not need to work for every node and game class. It is the responsibility of the user to instantiate a policy (and, as a consequence, a planning class)  that returns compatible data classes for the loss algorithm intended to be used. For example, if we want to update a policy function using the number of times a node has been visited during search (like AlphaZero), we should use a planning algorithm that returns a node class with the number of visits stored in each node, which is usually associated with best-first search algorithms.

%%%%%%%%%%%%%%%%%%%%%%%%%%%%%%%%%%%%%%%%%%%%%%%%%%%%%%%%%%%%%%%%%%%%%%
% IMPLEMENTATION
%%%%%%%%%%%%%%%%%%%%%%%%%%%%%%%%%%%%%%%%%%%%%%%%%%%%%%%%%%%%%%%%%%%%%%
\section{Implementation}
\label{sec:implementation}
In this section, we describe several environments and several module instances available in the proposed architecture and describe some examples of RL implementations using different combinations.  

\subsection{Environment}
The environment is modeled as a POMDP and the interface is similar to the openAI Gym environment \cite{OpenAIGym}. In each step of an episode, the agent executes an action, observing the next state and reward. Unlike the Gym environments, ours allows for multiple agents. Traditional problems where only one agent acts upon are conceptualized as a single agent problem. Another difference is the existence of illegal actions. The Gym environments have a fixed set of them in every state. The way they handle the actions chosen that do not make sense is by remaining in the same state and, often, returning a negative reward to disencourage these actions. The problem starts when we have more than one player. In Tictactoe, for example, players play in an alternating order; if the first player chooses a non-empty position on the board, under this solution, the player would still be the same in the next turn, which would break the consistency between player turns. The specific environments implemented are Tictactoe, Cartpole \cite{CartPole} and Minigrid \cite{minigrid}.
\par

\textbf{Cartpole}. In Cartpole, the agent has to balance a pole, which has one extremity fixated on a moving cart, by applying forces to the left and right of it. The environment has these two actions and  only one agent. The episode ends when the pole is more than 12 degrees from its original position or the cart is further than 2.4 units from the center. Optionally, we can limit the maximum number of steps. The observation is an array with four values: cart position, velocity, pole angle and angular velocity. \par
\textbf{Minigrid}. This environment is a $N\text{x}N$ grid. The agent starts in one of the squares and the objective is to reach a square goal. It has a direction and a position. The goal square is fixed, but we can optionally decide where the agent starts or choose for it to start in a random square every time the episode restarts. This environment is partially observable, being able to observe a grid of $7\text{x}7$ in each step. The observations are given in 7x7x3 arrays. The rewards are 1 if the goal is reached and 0 otherwise. There are three actions: turn left, right and move forward. All the actions are always legal and when the agent tries to continue past a wall, it simply does not move. Upon initializing this environment, we can limit the number of steps. \par
\textbf{Tictactoe}. It is a game played by two agents in a 3x3 grid. The two take turns filling an empty position in the grid. The first to fill three spaces in a row (horizontally, vertically or diagonally) wins. The observation is an array composed by two 3x3 arrays. The first has 1 in the positions where the current player has played and 0 otherwise, while the second has 1 in the positions the adversary has played and 0 otherwise. When initializing the environment, we can decide whether we want to play against an expert agent or we want to do \textit{self-play}, in which case the agents choose the actions for both players. The expert agent has three levels. The level 0 plays randomly. The next will play like the previous except if it sees moves that will block the third opponent piece in a row. In level 2, it also recognizes immediate winning moves. In self-play mode, the reward is always 0, except for a play that leads to victory, which is 1. In single agent mode it receives 1 if the action led to a victory, -1 to a defeat and 0 otherwise.

\subsection{Base Agent}
Just like in the environment module, we implement alternative algorithms in the agent's modules, which will allow us to compose different agents. However, there is a base idea behind all of these. To simplify, we call this common base the \textit{base agent}. \par
Our base agent learns the model of the environment as it interacts with it and, similarly to MuZero, the hidden states learned do not have a specific semantics. It will learn how to estimate the value of each state and the reward associated with each transition. The idea introduced, however, is to learn to predict the legal actions directly, so that it avoids exploring and conducting updates based on illegal parts of the tree during planning. MuZero avoids doing this using a policy function which sets the exploration of those nodes to 0, but this is unsuitable for algorithms that are non-best first or for those that are but estimate the value of the successors of the leaf expanded, instead of the leaf itself. By predicting the legal actions directly, our agent is much more compatible with different planning strategies. 

\subsection{Model}
This module should be easy to extend in order to accommodate any new functions needed by new search algorithms. The model should be used in the most efficient way possible, since deep learning computations can be time consuming. The efficiency, however, is dependent on the internal structure (the neural architectures used) and this needs to be easy to change. For instance, AlphaZero shares the same initial layers to calculate both the policy and value functions, which should not be calculated twice, but reused, in case we need both of them. To handle this, our implementation of the models has a \textit{query} method per type of input. For example, functions that receive a state share the same query method between them, but not with the functions that receive a state and an action as arguments. Each function has an attribute key associated with it that is passed to the respective query method when it needs to be calculated. In this way, if we want to calculate multiple functions (with the same input type), we make only one call to a query method, pass all the keys accordingly and the model decides internally how to calculate each one of the requested functions, reusing the necessary calculations for the highest efficiency. In practice, this is implemented using a hierarchical class structure that allows the Planning and Loss module to easily verify if the model passed to them supports every needed functions.\par

There are five relevant functions used by our base agent. The first, representation function, $h_\theta$, receives an observation and returns an encoded hidden state. The next two, state value function, $v_\theta$, and mask function, $m_\theta$ receive an encoded state and return, respectively, the estimated value for that state  and a predicted mask vector for its successors. Each index in the mask value represents whether the agent estimates the action to be legal or illegal, respectively giving values closer to 1 or to 0. The last two, reward, $r_\theta$, and next state, $g_\theta$, functions receive an encoded state and an action and return, respectively, the predicted reward associated with that transition and the next predicted state. \par
We implement one specific model (\textit{disjoint multilayer perceptron}) that supports the previously described five functions. Each one of them, internally, is computed by an independent multi-layer perceptron. \par

\subsection{Data}
\quad \textbf{Game submodule}. We implemented only one class that stores the information given by the environment during the episode: observations, actions, rewards, legal actions and players, and the data (nodes) returned by the planning algorithm.

\textbf{Node submodule}. Different types of search strategies might need different types of nodes. We implement three. The first one stores: a hidden state $s$; an improved state value $V(s)$; each successor's hidden state, $S(s, a)$, and their validity value $M(s, a)$, according to an action mask; the state’s player, $Pl(s)$; and the reward per transition, $R(s, a)$. The second node implemented is an extension
of the first. It is supposed to be used by best first algorithms and it stores, additionally, the number of visits in the tree, $N(s)$. Finally, the third node is an extension of the second one and it is designed to be used only by a Monte Carlo tree search. It stores a sum of state value estimations, $T(s)$, and it redefines the calculation of the improved state value to be the average of them:
\begin{equation} \small
	V(s) = \frac{T(s)}{N(s)}.
\end{equation}

\subsection{Planning}
A search algorithm will receive the current observation, player and action mask. After the search is done, it will return a (root) node associated with the current state of the environment. A tree structure resulting from the search algorithm is attached to this node, recursively through node successors. We will use interchangeably the term node and state (since the properties stored in the node refer to its state).\par
We implemented four search algorithms. They all use the five functions discussed before. An encoded state and a mask vector are predicted for every state in the tree. We define the \textit{action-value} as
\begin{equation} \small
Q(s^k,a) =
\left\{
\begin{array}{ll}
R(s^{k},a) + V(s^{k+1})  & \mbox{if } Pl(s^k) = Pl(s^{k+1}) \\
R(s^{k},a) - V(s^{k+1}) & \text{otherwise}
\end{array} ,
\right.
\end{equation}
where $s^{k+1}= S(s^k,a)$. When relevant, we use a superscript $k$ to denote that the state $s^k$ was created looking ahead $k$ steps. We can think of it as depth of the state in a hypothetical trajectory. The action-value can be seen as a 1-step bootstrapped value. We will also define the notion of masked action-value,
\begin{equation} \small
Q_M(s,a) = M(s,a) \cdot Q(s,a) + \Big(1- M(s,a)\Big) \cdot c_\text{penalty}
\label{masksucc1}
\end{equation}
For a perfect mask, we see that if the state is valid then the masked action-value is equal to the simple action-value. Otherwise, it is equal to the penalty constant, $c_{penalty}$, which penalizes state invalidity. Without the penalty, an invalid successor will have the value of zero, which might still be better than the alternative valid successors. An invalid state should never be chosen so, in theory, the constant $c_\text{penalty}$ should be minus infinity. However, in practice, the predicted masks are not perfect so we have to choose a milder value, such as a lower bound of the environment's reward function. The penalty term should only be used to choose an action, not to estimate the value of a state. \bigbreak

The four algorithms implemented start in the same way. They begin by creating the root node and the first encoded state using the representation function, $h_\theta$, based on the observation passed as argument. If the environment has two players, then its children nodes will be instantiated with the opposite player. After this first part, the algorithms start differing. \par
\textbf{Breadth First}. There are two breadth first algorithms: \textit{\bf minimax} and a variation of our own that we call \textit{\bf averaged minimax}. Both incrementally build a tree until depth $d$. In each iteration, they gather all the current leaf states and expand them until reaching the maximum depth, one depth level at a time (in a \textit{breadth first} manner). During a node's expansion, the expanded state's mask, the successor's encoded state and the respective transition reward are predicted. If the successor is at maximum depth, their state value is also predicted.
After the whole tree has been generated, the values of the leaf states are propagated backwards until the root. These two algorithms differ in the backup rule. For some state $s$, the update rule in the averaged minimax is given by
\begin{equation} \small \label{maskedAverage}
V(s) \leftarrow 
\frac
{\sum_{a\in\mathcal{A}}M(s,a)\cdot Q(s,a)}
{\sum_{a\in\mathcal{A}}M(s,a)} .
\end{equation}
For a perfect mask, we can see that the value of each state becomes the average of all the legal successor values. On the other hand, in minimax, the best successor is the one that maximizes the masked successor value, but, when updating the actual state value, we assume the successor to be completely valid:
\begin{equation} \small \label{maskedminimaxeq}
V(s) \leftarrow Q(s,\argmax_{a} Q_M(s,a)).
\end{equation}

\textbf{Best First}. The other two implemented algorithms are of type best first: \textit{\bf Monte Carlo tree search} (MCTS) and  \textit{\bf best first minimax} (BFMMS), both using a variation of the UCB formula (used in the UCT algorithm \cite{UCT}) to take into account the mask, resulting in what we call masked upper confidence bound  (MUCB). In each step of a best first iteration, the next action to be selected in a state, $s$, is given by
\begin{equation} \small \argmax_{a} \bigg(
Q_M(s,a)
+
c_{mucb} \cdot \sqrt{\frac{\log(N(s))}{N(S(s,a))+1}} \cdot M(s,a) \bigg) , 
\end{equation}
where $c_{mucb}$ is a constant that mediates the intensity of the exploration.

In an iteration, when expanding a leaf node, the states of its successors and their state values are estimated. The rewards and mask associated with these transitions are also predicted. The difference between MCTS and BFMMS is also in the backpropagation part. In BFMMS, the update rule is the same as in minimax (equation \ref{maskedminimaxeq}). In MCTS, the backpropagation will add, to each node, a path value, $G$, estimated after expanding the leaf node. The value added to the leaf after its expansion at depth $d$ is the masked average of its newly created successors
\begin{equation}  \small
G^d = 
\frac
{\sum_{a\in\mathcal{A}}M(s^d,a)\cdot Q(s^d,a)}
{\sum_{a\in\mathcal{A}}M(s^d,a)} .
\end{equation}
For every node in shallower depths after that, the added quantity is recursively given by
\begin{equation} \small
G^k =
\left\{
\begin{array}{ll}
R(s^{k},a^k) + G^{k+1}  & \mbox{if } Pl(s^{k}) = Pl(s^{k+1}) \\
R(s^{k},a^k) - G^{k+1} & \text{otherwise}
\end{array} ,
\right.
\end{equation}
where $0\leq k<d$.  In simpler words, $G$ accumulates the rewards as it backtracks. This path value is added to each node and their improved state value is given by the average of all of them
\begin{equation} \small
V(s^k) \leftarrow \frac{\sum_{i=0}^{N(s^k)}G_i^k}{N(s^k)} .
\end{equation}

\subsection{Policy}
In each step, the policy will use the planning algorithm to choose the next action. It will iterate with the environment until the episode is over, saving the planning and environment's data using the Game submodule. We implemented three policies. The first is an \textit{\bf $\epsilon$-value greedy}. After calling the planning algorithm and receiving the root node, it will choose with 1-$\epsilon$ probability the action corresponding to the highest action-value and choose a random one otherwise. The second and third policies are only applicable to best first algorithms since they base their choice on the number of visits. The second policy is \textit{\bf $\epsilon$-visit greedy}, which does the same as the first policy, but, instead of the action-value, it chooses based on the visits of each successor node. The third policy, \textit{\bf visit count}, is given by a visit count distribution just like in MuZero (equation \ref{AlphaZeroP}).

\subsection{Loss}
A loss class will receive a set of nodes and calculate a loss value as tensor that can be used to update the model. Each node has access to its game, which means it also has access to all the actions, observations and nodes of its episode. We implemented three loss algorithms. All of them use the same \textit{unrolling} process as MuZero and Value Prediction Networks to update the reward and mask functions: The state of the respective node is unrolled for $k$ steps (the next $k$ states are predicted using the actions taken during the real episode). The three losses differ in what target value to use to update the model. \par
\begin{itemize}
\item\textbf{Monte Carlo loss}. In this loss algorithm the target value, for a specific state, is the (discounted) sum of rewards from that state until the end of the trajectory.\par
\item\textbf{Offline Temporal Difference}. This is the one used in Muzero, where the target value is bootstrapped from the state $n$ steps away from the current ones, which is stored in the respective root node. \par
\item\textbf{Online Temporal Difference}. If the agent uses a replay buffer and reuses the same nodes for a long time, the value stored in them can easily become outdated. This loss is similar to the previous, but, instead of using the improved state value, $V(s_{t+n})$, it applies the representation function to the observation $n$ steps after the current one, $s_{t+n}=h_\theta(o_{t+n})$, and uses the value function to get a new value to bootstrap from, $v_\theta(s_{t+n})$.\par
\end{itemize}
If the unrolled states surpass the trajectory's size, they are modeled as absorbing states and the target value used with them is 0. \par

%%%%%%%%%%%%%%%%%%%%%%%%%%%%%%%%%%%%%%%%%%%%%%%%%%%%%%%%%%%%%%%%%%%%%%
% RESULTS
%%%%%%%%%%%%%%%%%%%%%%%%%%%%%%%%%%%%%%%%%%%%%%%%%%%%%%%%%%%%%%%%%%%%%%

\section{Experimental Evaluation}
\label{sec:resul}

In this section, we assess our implementation by creating different agents for the three environments implemented. For each environment, we look for the best combination of planning, loss and policy options, in this order. In the first step we compare planning algorithms and select the most promising one; in the second step we use the best planning method selected and compare different losses; finally, in the third step we keep the selected planning and loss options and compare policies. In the end, we should find a good agent for each environment, although no warranties exist that it is the optimum choice.	For each experiment, the agent plays a full episode, stores it and then conducts 10 updates, each with a batch size of 128 nodes. The results are shown as a function of training steps (the number of updates). In each experiment, the criterion to select the best algorithm is to choose the one with the highest mean score in the last step. As default, the loss used is the Monte Carlo one and the policy used is $\epsilon$-value greedy, whose parameters can be seen in each environment's second and third experiment, respectively.

\subsection{Cartpole}
\begin{figure}[h]
	\centering
	\includegraphics[width=0.9\linewidth]{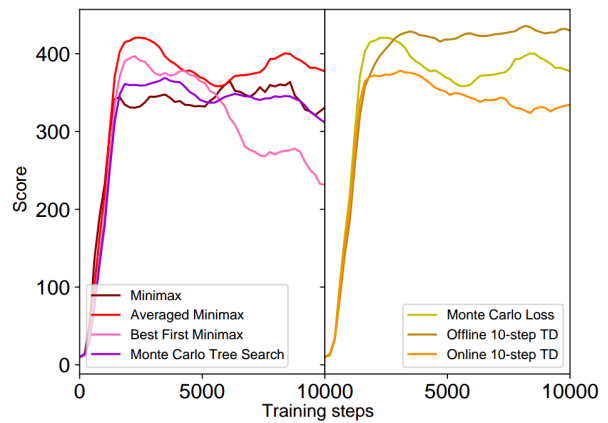}
	\caption{Results for Cartpole. From left to right: the experiences with Planning and Loss.}
	\label{fig:Cartpole}
\end{figure}

For Cartpole, we limit each episode to a maximum of 500 transitions and report, for each training step, the results as the mean of the last $100$ episodes. The Disjoint MLP has one layer, each with $80$ neurons, for every function. The encoded state used has size $12$. We use a uniform replay buffer with a maximum capacity of $10^5$ transitions. The masked penalty is $0$ as default. We repeated every experiment $10$ times, each for $10^4$ training steps.  The final results for every experiment for Cartpole can be seen in table \ref{table:Cartpole_results} and figure \ref{fig:Cartpole}. \par
\begin{table}
	\caption{Summary of the final results of the agents tested for Cartpole.}
	\centering
	\resizebox{1\linewidth}{!}{%
%	\begin{tabular}{ |p{2.5cm}|p{1.5cm}|p{2.5cm}||p{1cm}|p{1cm}|p{1cm}|p{1cm}|}
    \begin{tabular}{ |l|l|l||r|r|r|}
		\hline
		\multicolumn{6}{|c|}{Results for Cartpole} \\
		\hline
		Planning & Loss & Policy  & Max & Min & Mean  \\
		\hline
		Avg. Minimax & MC loss & $\epsilon$-value greedy    & 463.0   	&188.6	&377.4 	\\
		Avg. Minimax & ON TD   & $\epsilon$-value greedy    & 393.3   	&255.5	&334.4  \\
		Avg. Minimax & OFF TD  & $\epsilon$-value greedy     & \textbf{476.5}  	&\textbf{318.7} &\textbf{429.6}	 \\
		BFMMS 		 & MC loss & $\epsilon$-value greedy     & 450.7    	&119.5	&231.6 	 \\
		Minimax 	 & MC loss & $\epsilon$-value greedy    & 401.6    	&258.0	&331.1 	 \\
		MCTS 		 & MC loss & $\epsilon$-value greedy    & 473.1    	&142.6	&311.5 	\\
		\hline
	\end{tabular}
	}
	\label{table:Cartpole_results}
\end{table}

\begin{enumerate}
\item \textbf{Planning}. The best first algorithms use $16$ iterations and the breadth first ones explore a maximum depth of $4$. The best algorithm was the averaged minimax, which is used in the next experiments. \par
\item \textbf{Loss}. The loss algorithms use $10$ unroll steps and a gamma discount of $0.95$. For the temporal difference losses, the bootstrapping step is $10$. The best loss was the offline temporal difference.\par 
\item \textbf{Policy}. There is only one policy that is compatible with breadth first algorithms. So, the averaged minimax with offline temporal difference and $\epsilon$-value greedy is the best algorithm combination.
\end{enumerate}

\subsection{Minigrid}
\begin{figure}[h]
	\centering
	\includegraphics[width=1\linewidth]{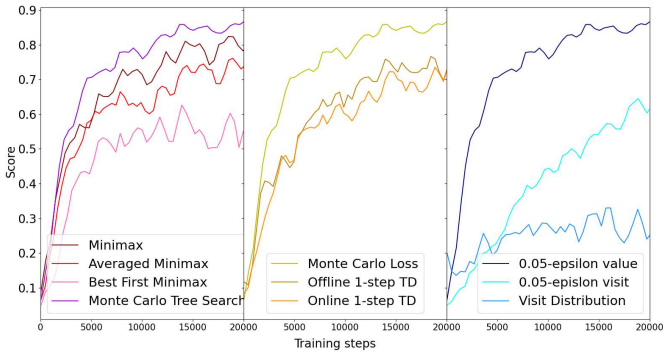}
	\caption{Results for Minigrid. From left to right: the experiences with Planning, Loss for averaged Minimax, Loss for MCTS and Policy.}
	\label{fig:Minigrid}
\end{figure}

We use a $6\text{x}6$ Minigrid environment. Each episode starts at a random position and is limited to a maximum of $15$ transitions. The Disjoint MLP has one layer of $100$ neurons per function and the hidden state has size $15$. We use a prioritized replay buffer with a maximum capacity of $25000$ transitions. The masked penalty used is $0$. We repeated every experiment $10$ times, each for $20000$ training steps. The results are given, per training step, as the mean score of the last $100$ episodes. The final results for every experiment for Minigrid can be seen in table \ref{table:Minigrid_results} and figure \ref{fig:Minigrid}.\par

\begin{table}
	\caption{Summary of the final results of the agents tested for Minigrid.}
	\centering
	\resizebox{1\linewidth}{!}{%
%	\begin{tabular}{ |p{2.5cm}|p{1.5cm}|p{2.5cm}||p{1cm}|p{1cm}|p{1cm}|p{1cm}|}
        \begin{tabular}{ |l|l|l||r|r|r|}
		\hline
		\multicolumn{6}{|c|}{Results for Minigrid} \\
		\hline
		Planning & Loss & Policy  & Max & Min & Mean  \\
		\hline
		Avg. Minimax 	& MC loss & $\epsilon$-value greedy    &0.90  	&0.60	&0.76 	\\
		BFMMS 	& MC loss & $\epsilon$-value greedy    &0.63  	&0.41	&0.54 	\\
		MINIMAX 	& MC loss & $\epsilon$-value greedy    &0.91  	&0.71	&0.79 	 \\
		MCTS 	& MC loss & $\epsilon$-value greedy    &\textbf{0.99} 	&\textbf{0.76}	&\textbf{0.87}  \\
		MCTS 	& MC loss & $\epsilon$-visit greedy    &0.87 	&0.00	&0.62 	 \\
		MCTS 	& MC loss & Visit dist.   &0.33 	&0.22	&0.26 	 \\
		MCTS 	& OFF TD & $\epsilon$-value greedy    &0.91	&0.63	&0.73 	 \\
		MCTS 	& ON TD & $\epsilon$-value greedy    &0.86 	&0.52	&0.73 	 \\
		\hline
	\end{tabular}
	}
	\label{table:Minigrid_results}
\end{table}

\begin{enumerate}
\item \textbf{Planning}. The number of iterations used by the best first algorithms is $27$ and the breadth first ones use a depth of $3$. MCTS was the best planning algorithm.\par
\item \textbf{Loss}. We use $7$ unroll steps and a gamma discount of $0.99$ in every loss algorithm. For the temporal difference losses, we use a bootstrapping step of $1$. The best loss was the Monte Carlo loss.\par
\item \textbf{Policy}. We use $5\%$ exploration for the $\epsilon$-greedy strategies and a temperature of $1$ for the visit distribution policy. The best agent found was MCTS with Monte Carlo loss and $\epsilon$-value greedy.
\end{enumerate}

\subsection{Tictactoe}
\begin{figure}[h]
	\centering
	\includegraphics[width=1\linewidth]{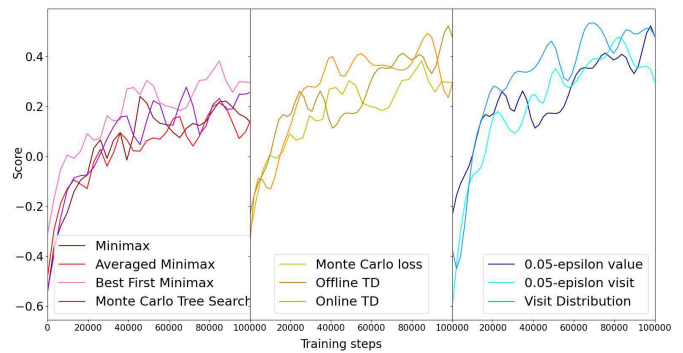}
	\caption{Results for Tictactoe. From left to right: the experiences with Planning, Loss for averaged Minimax, Loss for MCTS and Policy.}
	\label{fig:Tictactoe}
\end{figure}

The agents are trained in this environment by doing self-play. Every $5000$ training steps, we play $100$ games against an expert agent (level 1 in our implementation) and report the mean score. When playing against the expert we remove exploration from the policy. The Disjoint MLP has one layer of $100$ neurons in every function, except in the next state function, which has $200$. The hidden state has a size of $32$. We use a prioritized replay buffer with a maximum capacity of $25000$ transitions. The masked penalty is $-1$. Each experiment is ran for $10^5$ training steps and repeated $5$ times. The final results for every experiment for Tictactoe can be seen in table \ref{table:Tictactoe_results} and figure \ref{fig:Tictactoe}. \par

\begin{table}
	\caption{Summary of the final results of the agents tested for Tictactoe}
	\centering
	\resizebox{1\linewidth}{!}{%
%	\begin{tabular}{ |p{2.5cm}|p{1.5cm}|p{2.5cm}||p{1cm}|p{1cm}|p{1cm}|p{1cm}|}
        \begin{tabular}{ |l|l|l||r|r|r|}
		\hline
		\multicolumn{6}{|c|}{Results for Tictactoe} \\
		\hline
		Planning & Loss & Policy  & Max & Min & Mean  \\
		\hline
		Avg. Minimax 	& MC loss & $\epsilon$-value greedy    &0.28  	&0.01	&0.14 	 \\
		BFMMS 	& MC loss & $\epsilon$-value greedy    &0.48  	&-0.02	&0.29 	 \\
		BFMMS 	& OFF TD & $\epsilon$-value greedy    &\textbf{0.54} 	&0.39	&\textbf{0.48} 	 \\
		BFMMS 	& OFF TD & $\epsilon$-visit greedy    &0.41 	&0.14	&0.30 	 \\
		BFMMS 	& OFF TD & Visit dist.    &0.53  	&\textbf{0.45}	&\textbf{0.48} 	 \\
		BFMMS 	& ON TD & $\epsilon$-value greedy    &0.38  	&0.23	&0.29 	 \\
		Minimax 	& MC loss & $\epsilon$-value greedy     &0.21  	&0.12	&0.14 	\\
		MCTS 	& MC loss & $\epsilon$-value greedy     &0.35  	&0.02	&0.25 	 \\
		\hline
	\end{tabular}
}
	\label{table:Tictactoe_results}
\end{table}

\begin{enumerate}
\item \textbf{Planning}.  The number of iterations used by the best first algorithms is $25$ and the breadth first ones use a depth of $2$. The best algorithm was the BFMMS. \par
\item \textbf{Loss}. We use $5$ unroll steps and a gamma discount of $0.99$. For the temporal difference losses, we use a bootstrapping step of $1$. The offline temporal difference was the best loss.\par
\item \textbf{Policy}. The policies use $5\%$ exploration for the $\epsilon$-greedy strategies and a temperature of $1$ for the visit distribution policy, when training, and choose always the best action when testing against the expert. There is a tie between two agents, which use BFMMS and offline temporal difference, but differ in policy. One uses the $\epsilon$-value greedy and the other the visit distribution. The combination that uses visit distribution as policy, however, is, throughout the experiment, consistently better than the one with $\epsilon$-value greedy (fig. \ref{fig:Tictactoe})
\end{enumerate}

\subsection{Discussion}
Every problem is different and the results captured for these do not automatically guarantee the same for any other. However, these have distinctive characteristics of their own, making them good \textit{testbeds} to see if our implementation works. Cartpole has a strong reward signal, but is very long; as opposed to Minigrid, which has a sparse reward signal but short episodes. These last two do not have illegal actions, which make them suitable to see if our implementation works in \textit{common} environments. Tictactoe, on the other hand, can be seen as the final challenge since it requires a good use of the mask learned and it uses self-play used during training, a technique known to be challenging in reinforcement learning.  In all of these, the best agents found achieved very good results, reaching almost the maximum score in Cartpole and Minigrid. For Tictactoe, the expert is good at defending against obvious plays and it is not good at attacking, so it is expected that a very good agent will win frequently when it starts and tie when the opponent does, leading to a score near $0.5$. Our best agent gets very close to this value. The idea of predicting a mask directly and the algorithm adaptations described in this work seem to function well. \par
The most surprising result is the best planning algorithm in Cartpole being the averaged minimax. This algorithm was created by us in this work based on the very \textit{naive} idea of averaging over all successors in each node. Out of all four algorithms, it is the only one that does not theoretically converge to the minimax function. Still, it was better than all the others in Cartpole; achieved a better result than BFMMS in Minigrid and had the same result as minimax in Tictactoe.\par
Regarding the new loss, the online temporal difference, it was not particularly better than the other two, but it was still fairly consistent, achieving better results than Monte Carlo loss in Cartpole and the same in Minigrid.\par
The most important observation, however, is that, for each component, different algorithms were better according to the environment. For example, the BFMMS, that was the worst in Cartpole and Minigrid, was the best in Tictactoe. Similarly, the visit distribution policy that got poor results in the initial two, achieved the maximum mean score in the third environment. This motivates the existence of flexible, modular, architectures that facilitate the assessement of different choices for policy, loss and planning approaches. The existence of architectures that make such assessements easier suggests the possibility of improving state of the art MBRL agents, like Muzero, in certain environments, by modifying their components and finding more suitable algorithm combinations.

%%%%%%%%%%%%%%%%%%%%%%%%%%%%%%%%%%%%%%%%%%%%%%%%%%%%%%%%%%%%%%%%%%%%%%
% CONCLUSIONS
%%%%%%%%%%%%%%%%%%%%%%%%%%%%%%%%%%%%%%%%%%%%%%%%%%%%%%%%%%%%%%%%%%%%%%

\section{Conclusions and Future Work}
\label{sec:concl}

In this work, we presented a modular architecture that identifies separate relevant components for MBRL agents that use planning algorithms. Then we provided our own specific implementation of this architecture, capable of searching in domains with a variable set of actions in each state, by learning an action mask directly. We implemented several search algorithms, adapting them to the mask. One of them, called averaged minimax, was proposed in this work and yield surprisingly good results. Different loss methods, policies and environments were also implemented. Lastly, we experimented this implementation by creating different agents and demonstrated that it works and achieves good results for the environments implemented. The experimentation done seemed to indicate that the best algorithm combination in an agent is problem dependent, motivating the need for tools that facilitate the implementation and variation of the different parts of an agent, like the ones introduced in this work.

\subsection{Future Work}
This work was motivated by the need for tools that allow  extension, so it is only natural that we have some suggestions about how to extend it. The most obvious suggestion is to implement AlphaZero and MuZero and study if, by varying some of their parts, it may be possible to bring the hardware requirements down. If the hardware resources allow it, a comparison, using more challenging environments, between these agents and our implementation might be interesting.  \par
Some other ideas are, for instance, teaching the model to predict the observations perfectly; using every node in the search tree as data \cite{GameTreeBootstrapping}; add loss methods based on the TD($\lambda$) algorithm \cite{SuttonAndBarto}; implement our base agent for the case of non-deterministic environments (which is also a suggestion described in the paper of Muzero \cite{Muzero}); add search algorithms like A*\cite{Astar}, K-BFS \cite{KBFS}, IDA* \cite{IDAstar}, B* \cite{Bstar}, among other. \par
We do not expect our work to accommodate all possible MBRL algorithms, as that would be too ambitious, but it should be able to enable researchers to implement a very diverse range of ideas that are common in reinforcement learning research.\par

%%%%%%%%%%%%%%%%%%%%%%%%%%%%%%%%%%%%%%%%%%%%%%%%%%%%%%%%%%%%%%%%%%%%%%
% ACKNOWLEDGMENTS
%%%%%%%%%%%%%%%%%%%%%%%%%%%%%%%%%%%%%%%%%%%%%%%%%%%%%%%%%%%%%%%%%%%%%%

%%%%%%%%%%%%%%%%%%%%%%%%%%%%%%%%%%%%%%%%%%%%%%%%%%%%%%%%%%%%%%%%%%%%%%
% REFERENCES
%%%%%%%%%%%%%%%%%%%%%%%%%%%%%%%%%%%%%%%%%%%%%%%%%%%%%%%%%%%%%%%%%%%%%%

% Produces the bibliography section when processed by BibTeX
%
% Bibliography style
% > entries ordered alphabetically
%\bibliographystyle{plain}
% > unsorted with entries appearing in the order in which the citations appear.
%\bibliographystyle{unsrt}
% > entries ordered alphabetically, with first names and names of journals and months abbreviated
\bibliographystyle{abbrv}
% > entries ordered alphabetically, with reference markers based on authors' initials and publication year
%\bibliographystyle{alpha}

% External bibliography database file in the BibTeX format (ExtendedAbstract_ref_db.bib)
\bibliography{ExtendedAbstract.bib}

%%%%%%%%%%%%%%%%%%%%%%%%%%%%%%%%%%%%%%%%%%%%%%%%%%%%%%%%%%%%%%%%%%%%%%
\end{document}